\documentclass{article}

\usepackage{iclr2027_conference,times}
\usepackage[T1]{fontenc}
\usepackage[utf8]{inputenc}
\usepackage{microtype}
\usepackage{amsmath,amssymb}
\usepackage{graphicx}
\usepackage{booktabs}
\usepackage{multirow}
\usepackage{placeins}
\usepackage{xcolor}
\usepackage{tikz}
\usetikzlibrary{positioning, arrows.meta}
\usepackage[hidelinks]{hyperref}
\usepackage{url}
\usepackage[capitalize]{cleveref}
\crefname{subsection}{Section}{Sections}
\Crefname{subsection}{Section}{Sections}
\crefname{subsubsection}{Section}{Sections}
\Crefname{subsubsection}{Section}{Sections}
\crefname{appsec}{Appendix}{Appendices}
\Crefname{appsec}{Appendix}{Appendices}
\creflabelformat{equation}{#2#1#3}

\title{Hierarchical Token Merging with\\Recursive Autoencoders for\\Unsupervised Grammar Induction}

\author{Murat Guney Kemal \\
Department of Computer Science\\
Durham University\\
\texttt{murat.guney.kemal@outlook.com}
\And
Robert Lieck \\
Department of Computer Science\\
Durham University\\
\texttt{robert.lieck@durham.ac.uk}
}

\iclrfinalcopy

\begin{document}

\maketitle
\lhead{}
\renewcommand{\headrulewidth}{0pt}

\begin{abstract}
Hierarchical abstraction is a fundamental principle across many data modalities, enabling complex sequences to be represented compositionally at multiple levels. Learning such hierarchies directly from data, without explicit structural supervision, remains a central challenge in representation learning.
We investigate whether meaningful hierarchical structure can emerge from compression alone. While autoencoders learn compressed representations through reconstruction, this objective does not necessarily favour meaningful compositional structure. We study this challenge through unsupervised grammar induction, where the underlying hierarchy can be evaluated explicitly.
We introduce a hierarchical token-merging autoencoder that recursively combines adjacent tokens into increasingly abstract representations, compressing entire sequences through a variational information bottleneck. The model jointly learns representations and hierarchical structure by marginalising over possible binary trees using dynamic programming. To address training instability, we combine independently trained models through unsupervised consensus and distillation.
Experiments on a synthetic, ambiguous grammar demonstrate the emergence of meaningful hierarchical structure without structural supervision. The resulting model achieves 93\% grammatical parse-validity and 93\% exact reconstruction on held-out sequences, compared with 42\% parse-validity for the strongest trivial baseline. These findings establish a proof of concept for hierarchical token merging as a mechanism for unsupervised structure discovery and abstraction learning.

\end{abstract}

\section{Introduction}

Language is compositional, and so are many other kinds of sequence. Small units combine into larger ones hierarchically, and
those into larger ones again. Recovering that structure without annotation would be useful, because the
treebanks that supervised parsers rely on are expensive to build and specific to one language and domain, and
because whether grammatical structure can emerge with no supervision at all is a question of independent
scientific interest. The
question we study is whether a model can find constituent structure in raw token sequences, with no labelled
trees to learn from, with an information bottleneck as the only pressure towards structure.

This is hard for two reasons. First, a plain reconstruction objective is grammar-agnostic. An autoencoder that only
has to reproduce its input is rewarded for whatever structure is easiest to reconstruct, which may not be the
grammatical one. Second, the structure is a discrete object from a large space, since a sentence of $n$ tokens has
exponentially many binary trees; learning the representations and searching that space at the same time is
unstable. Models that induce latent trees often do well on their task without recovering structure that
matches human annotations \citep{williams2018latent}, so reconstructing or predicting well is not the same as
parsing well.

We avoid choosing a tree at all. Instead of deciding which pair to merge, the model scores every binary tree
at once with an inside chart, a dynamic program that sums over all trees in cubic time, and reads a single
parse off the chart only at test time. The merge stays a hard decision, made differentiable with a
straight-through estimator. The pressure towards structure comes from the variational bottleneck. We hypothesise that when a whole sentence has to
pass through one small vector, merging it along its grammatical constituents has the lowest reconstruction and
compression cost, so the model comes to prefer grammatical structure because it compresses well, not because we
supervise it. It is never shown
a single tree, so it becomes a parser implicitly, as a by-product of learning to compress.

This leads to our research question. \emph{Can a token-merging autoencoder induce grammatical structure from
raw tokens with no supervision, by scoring all possible trees at once, and if that induction works but is unstable, can
the instability be removed without reintroducing supervision?}

We make the following contributions.
\begin{itemize}
\item We show that a token-merging autoencoder can induce grammatical structure from raw tokens with no
supervision, and from a deliberately minimal self-supervision (compression through a variational
bottleneck alone, with no outside pass and no reconstruction of words from their context), by
marginalising over trees rather than choosing a merge. In the best runs it recovers the grammar
with held-out parse-validity of 1.00.
\item We show that a single model is unreliable. Its induction is what we call a seed lottery. With
identical data and configuration, held-out parse-validity ranges from 0.01 to 1.00 depending only on the random
seed, and even the best-parsing models reconstruct
poorly, because the tight bottleneck that drives them to find structure also costs them the ability to
reproduce their input.
\item We find that combining many models at the level of individual constituents (span-based minimum Bayes
risk, span-MBR) beats combining whole trees, which matters on ambiguous data where good models disagree over equally
valid parses. \citet{shayegh2024ensemble} use the same combination for ensembles of unsupervised parsers;
we show it remains effective when the members are runs of one model whose quality ranges from 0.01 to 1.00.
\item To make the method robust despite that lottery, we build a pipeline that trains many models, selects the
mutually consistent ones, combines them and distils the result into a single model that reaches a mean held-out
parse-validity of 0.93 and also reconstructs its input, which no single stage does alone.
\item We report a seven-pool robustness study, varying both the random seed and the data split, showing that
unsupervised selection mainly raises the weakest pools rather than improving every pool.
\item We find that, under the training schedules we tested, a single model does not reach both grammar and
reconstruction, because the two objectives compete for its capacity, so in our setting the two-stage separation
of induction from reconstruction is necessary rather than merely convenient.
\item We find that the two stages learn the grammar differently. The chart-trained base model, under the
compression bottleneck, organises its vocabulary by syntactic category as it discovers constituents, whereas
the distilled student reaches the same parse accuracy without clustering its word embeddings by category, so category structure is a
by-product of how structure is induced, not a prerequisite for parsing.
\end{itemize}

Two caveats bound these claims, and \Cref{sec:limitations} takes up both in full. The reported results are a
distribution over independently seeded pools rather than a single number, and one hyperparameter, the number
of models retained, was chosen with knowledge of gold-scored results. Gold trees, by contrast, are used only to
evaluate the final model, never in training, model selection, or the combination of the models' parses (the
consensus).

\Cref{sec:related} places the method among grammar induction, chart-based marginalisation and neural latent-tree
models. \Cref{sec:method} describes the autoencoder, the chart, and the pipeline that turns many models into one.
\Cref{sec:experiments} presents the experiments, the aggregation choice, the seed lottery, the combiner, the seven-pool
result and what the latent space encodes, and closes with the limitations and future work. \Cref{sec:conclusion} answers the research
question.

\section{Related Work}
\label{sec:related}

We survey six literatures that bear on the method. They are unsupervised grammar induction, chart-based
marginalisation over trees, neural latent-tree induction, the structured-autoencoder and discrete-latent
machinery the model is built from, the separate token-merging line in the efficiency literature that shares
our name, and the consensus-decoding and distillation methods our pipeline uses to combine models. We took
two papers as anchors, the DIORA inside-outside autoencoder and the Recursive Bayesian Network framework, and
followed citations forward and backward from each, supplementing with keyword searches for the individual
components (variational bottleneck, straight-through estimation, minimum Bayes risk decoding, knowledge
distillation).

\subsection{Grammars, constituency and structure induction}
Grammars describe how sequences decompose into nested constituents. A context-free grammar rewrites symbols
into strings of symbols, and a probabilistic context-free grammar (PCFG) attaches a probability to each rule,
so a sentence has a distribution over parse trees rather than one. The main difficulty for evaluation is ambiguity. A
single surface (token sequence) can have several valid parses, so a model may be right and still not match the one tree a
corpus happens to record. This shapes how we evaluate, and we come back to it in \Cref{sec:setup}. Unsupervised
induction of such grammars has a long history. The inside-outside algorithm estimates a PCFG by
expectation-maximisation \citep{baker1979inside}; the constituent-context model induces bracketings from the
distributions of spans and their contexts \citep{klein2004ccm}; and neural and compound PCFGs parameterise the
rule probabilities with neural networks, which induces better grammars than classical PCFG estimation
\citep{kim2019compound, yang2021pcfg}.
All of these learn an explicit symbolic grammar. Our model learns no rules. Structure is a by-product of
compressing the sequence, and there is no grammar to read off, only a tree per sentence.

\subsection{Marginalising over trees with CKY, the inside algorithm and Recursive Bayesian Networks}
\label{sec:rbn}
To score structure without choosing it we use a chart. The Cocke-Younger-Kasami recogniser fills a table
over spans and, for each span, combines the analyses of its sub-spans \citep{younger1967cyk}; the inside
algorithm makes this probabilistic, summing over all derivations of a span in cubic time \citep{baker1979inside}.
That sum is a differentiable function of the local scores, so a chart can be embedded in a neural model and be
trained by gradient descent. Recursive Bayesian Networks (RBNs) generalise the idea \citep{lieck2021rbn}. They
unify PCFGs, which have discrete symbols and tree structure, with dynamic Bayesian networks, which have
continuous latents and chain structure, and they extend inside and outside probabilities to continuous latent
variables. Our chart is an RBN inside pass whose latent at each span is a neural embedding. The framework also
suggests an explanation for the seed instability we report in \Cref{sec:lottery}. Marginalising over
exponentially many structures turns the objective into a mixture with one component per structure, which is
non-convex, so gradient descent is not guaranteed to reach the global optimum, and optimising the
representations and the structure in turn (as expectation-maximisation does) is not the same as optimising them
together, as our model does. We interpret the seed dependence in \Cref{sec:lottery} as a consequence of this
non-convexity.

\subsection{Neural latent-tree induction}
The closest prior work is DIORA, an unsupervised inside-outside recursive autoencoder \citep{drozdov2019diora}.
DIORA composes span representations with a soft, compatibility-weighted average on the inside pass, runs an
outside pass that reconstructs each word from its context, and reads a tree off the chart at test time. On the
full Wall Street Journal corpus it reaches an unlabelled constituency F1 of 55.7 on average against a
right-branching baseline of 16.5, and it varies little across seeds (about $\pm 0.5$ F1). Our chart differs
from it in four ways. It is inside-only, so we drop the outside pass. It aggregates a span with a hard
straight-through choice instead of a soft blend, so a span is one analysis and not a mixture of all of them. It
reconstructs the whole sentence from a single root vector through a variational bottleneck, instead of
predicting each word from its local context, so it has an explicit information bottleneck where DIORA has none.
And its objective, which must reconstruct the whole sentence from a single bottlenecked vector, is not
seed-stable where DIORA is; the ensemble and distillation stages of our method are designed to compensate for
that instability (\Cref{sec:lottery}). Other neural models induce a latent tree jointly with a downstream
objective \citep{yogatama2017rlspinn, choi2018gumbeltree, havrylov2019cooperative}, and syntax-aware sequence
models bias recurrent networks toward hierarchy without an explicit chart \citep{shen2018prpn, shen2019ordered}.
\citet{williams2018latent} show that latent-tree learners can do well on their end task while the trees they
induce do not match meaningful structure. Accordingly, our evaluation
reports grammatical validity rather than claiming linguistic correctness.

\subsection{Structured autoencoders and discrete latent variables}
Our model combines several standard components. Recursive autoencoders learn representations of tree nodes by
encoding and decoding their children \citep{socher2011rae}, and the recursive neural network that first built
trees for scenes and language did so by greedily merging the adjacent pair with the best composition score
\citep{socher2011parsing}. We keep the recursive merge of adjacent pairs but replace the greedy score with
marginalisation over all trees. The bottleneck is a variational autoencoder \citep{kingma2014vae}; its
$\beta$-weighted form sets the weight of the penalty on the latent's divergence from its prior, and hence how
much information the latent retains \citep{higgins2017beta}; we rely on this penalty to favour structured
encodings. To keep the merge a hard decision while still training by gradient descent
we use continuous relaxations of discrete variables, the Gumbel-softmax or Concrete distribution
\citep{jang2017gumbel, maddison2017concrete}, with the straight-through estimator that passes a hard sample
forward and a soft gradient backward \citep{bengio2013estimating}. We reuse these components unchanged. What
is new is not any one of them but their combination, a chart-based merge autoencoder in which a compression
bottleneck, rather than any label, is what drives the structure.

\subsection{Token merging in the efficiency literature}
The name \emph{token merging} also appears in a different literature, where tokens are merged to reduce the
computational cost of transformers, not to recover structure. Spectrum-preserving merging \citep{tran2024pitome} and
agglomerative token clustering \citep{haurum2024agglomerative} are recent examples, and one method,
dynamic hierarchical token merging for vision transformers \citep{haroun2024dynamic}, shares the name of our
method.\footnote{That method merges visual tokens to speed up a vision transformer and does
not address grammar induction; we merge tokens to recover constituent structure.} Closer to our
method, a line of work learns where to split a sequence end-to-end, as in dynamic chunking for hierarchical sequence modelling
\citep{hwang2025dynamic}; like our model, it learns the hierarchy end-to-end, but its objective is language
modelling rather than structure recovery.

\subsection{Consensus decoding and distillation}
The ensemble stage of our method builds on two ideas. Minimum Bayes risk decoding returns not the single
highest-scoring structure but the one that minimises expected loss under a distribution over structures; for
trees that means maximising the expected number of correct constituents \citep{goodman1996parsing}, which is why combining candidate
parses span by span behaves differently from voting over whole trees. A span-level combination can return a
tree that no single model proposed, whereas a whole-tree vote cannot. Ensembling independent models reduces
variance, and knowledge distillation compresses an ensemble into one model \citep{hinton2015distilling}. We use
both, with one difference. Our ensemble is a set of independently seeded parsers combined once, offline, into a
single set of silver trees (machine-produced trees used as training targets in place of gold annotations), not
an ensemble queried at test time.

The closest prior work on this side of the method is \citet{shayegh2024ensemble}, who combine an ensemble of
different unsupervised parsers by searching, for each sentence, for the binary tree that shares the most
constituents with the members' trees, and then distil the result into a single student. Our consensus is the
same construction. What differs is the ensemble. Ours consists of repeated runs of one model whose quality ranges
from failing to perfect, and we choose which members to keep without supervision.

\subsection{Summary and gap}
In our work we build a token-merging autoencoder that induces structure from a compression bottleneck alone,
together with a CKY inside pass that lets it score every tree instead of choosing one. This induction
succeeds in some runs but varies strongly across seeds, so a single model is a seed lottery. To turn that into one reliable parser we combine
many models and distil the result into one, an approach also taken by \citet{shayegh2024ensemble}. The rest of the paper describes the method and reports both what it does and
where it is fragile.

\section{Method}
\label{sec:method}

\subsection{Overview}
The task is to recover the constituent structure of a sentence without being shown any structure during
training. We work with sentences of between two and eight tokens drawn from a small probabilistic grammar in
which many sentences have more than one valid parse. Bounding the length keeps the number of possible binary
trees small enough to enumerate exhaustively at evaluation time. \Cref{sec:setup} describes the data in full.

\begin{figure}[t]
\centering
\begin{tikzpicture}[
  >=Latex,
  stage/.style={draw, rounded corners, align=center, text width=3.3cm, minimum height=2.4cm, inner sep=5pt,
                font=\small}
]
\node[stage] (s1) {\textbf{Stage 1}\\[2pt]Train $N$ base models; each parses by scoring all trees; freeze
each};
\node[stage, right=1.1cm of s1] (s2) {\textbf{Stage 2}\\[2pt]Select the mutually consistent third; combine
parses at the constituent level $\rightarrow$ silver treebank};
\node[stage, right=1.1cm of s2] (s3) {\textbf{Stage 3}\\[2pt]Distil one model on the silver treebank;
evaluate on held-out data};
\draw[->, thick] (s1) -- (s2);
\draw[->, thick] (s2) -- (s3);
\end{tikzpicture}
\caption{The three-stage pipeline. The same architecture is used as the base model in Stage 1 and as the
distilled model in Stage 3; only Stage 1 uses the chart, while Stage 3 trains on the fixed silver trees.}
\label{fig:pipeline}
\end{figure}
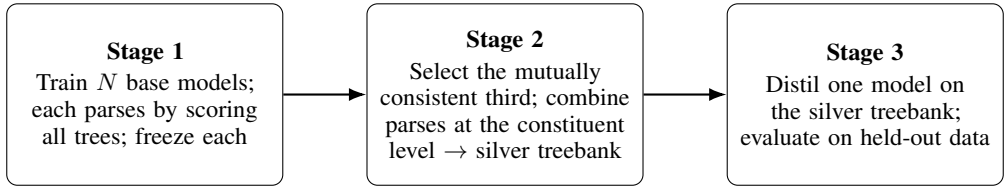

The pipeline uses a single architecture for two different jobs. That architecture, which we call the base
model, is a variational autoencoder built from four networks, a lifter, a merger, an unmerger and an
unlifter. \Cref{fig:pipeline} shows how the three stages fit together.

In the first stage we train $N$ independently initialised base models and freeze each one when training ends.
Its learning rate decays to near zero by step 2{,}500, after which its weights are fixed (\Cref{sec:baseobj}).
Every base model learns to compress a sentence into a single vector by repeatedly merging adjacent pairs,
choosing the merge order for itself. Trained with a tight bottleneck (a Kullback--Leibler (KL) weight of $\beta = 0.04$, \Cref{tab:config}), these models become good parsers but
poor reconstructors. They recover the grammar while losing the ability to reproduce the input tokens. Their
quality also varies sharply between runs, even though every model sees identical data and differs only in its
random initialisation and the order in which it receives the training examples. That variability is why we
train many models rather than one (\Cref{sec:lottery}).

In the second stage every frozen model parses the training sentences. We keep the third whose parses are most
often identical to those of the other models (peer agreement, \Cref{sec:select}), on the assumption that models which have found the grammar tend to agree with one
another while failed models disagree in idiosyncratic ways. The retained models then vote on individual
constituents rather than on whole trees, and a dynamic program assembles the best-supported constituents into
one tree per sentence. The result is a single silver treebank.

In the third stage we train one fresh base model on the silver treebank. The structure is now fixed, so the
model no longer has to discover how to bracket a sentence and can instead devote its capacity to the merge and
unmerge operations. This distilled model uses a weaker bottleneck than the others ($\beta = 0.002$ and a larger latent) and is the one we evaluate
on held-out sentences.

\subsection{The token-merging autoencoder}
We designed the base model as a recursive $\beta$-VAE whose decoder is guided by the tree along which the
encoder has merged. The model has four neural networks, a lifter, a merger, an unmerger and an unlifter. The
lifter maps tokens into the embedding space. We then apply a single merger recursively, first to the leaf
tokens and then to the intermediate constituent representations it produces, and this is the encoder.
Similarly we apply a single unmerger recursively to the root latent, and this is the decoder. The
unlifter maps the recovered latents to a distribution over the vocabulary, from which the output tokens are
predicted.

While training the base models we score all trees with the chart (\Cref{sec:chart}) rather than committing to a
single one, so every split of every span receives a gradient, rather than only the splits of one chosen tree. We kept the
base model deliberately simple, so that we could test whether the architecture induces grammar at all rather
than to maximise capacity; a small model also makes ablations easier to interpret.

\subsubsection{The four networks}
The lifter maps each token to a latent and is stochastic, with two embedding tables giving a mean and a
log-variance and the latent sampled by the reparameterisation trick, so a word is a distribution rather than a
point. The merger is the encoder step, a small multilayer perceptron that takes a left and a right latent and
outputs the mean and log-variance of their parent, a latent with the same dimension and prior as a leaf latent, so one merger runs at
any depth of the tree. The unmerger is the decoder step, a deterministic network trained to approximately invert the
merger, mapping a parent back to a predicted left and right child. The squared error between its predictions and the actual
children measures how invertible a merge is, and this same quantity scores the candidate splits in the chart
(\Cref{sec:chart}). The unlifter is a single linear layer from a latent to scores over the vocabulary, kept
deliberately weak so the model cannot use a powerful decoder to compensate for a poorly chosen structure. The
full architecture is shown in \Cref{app:arch}.

\subsubsection{Bottleneck and training on a fixed tree}
The forward pass on a given tree has two parts. We first map every token to a latent with the lifter, then
apply the merger bottom-up along the tree until a single root latent remains. Decoding works back down the
same tree through the unmerger, and finally the unlifter turns each recovered leaf latent into a distribution
over the vocabulary.

The decoder starts from the root latent alone and is given none of the encoder's intermediate latents. This
forces the model to compress, and our assumption is that the lowest-cost way to encode a sentence in one vector
is to merge it along its grammatical constituents. If we carried the encoder's
intermediate latents across, the model could reconstruct without compressing and there would be no pressure to
find structure.

We regularise both leaf latents and merged latents toward the same $N(0,I)$ prior. We chose a shared prior so
that the KL is measured on the same scale at every level of the tree, and so that a single merger can operate
at any level; in the objective below this KL is weighted by $\beta$.

On a given tree we minimise three terms:
\begin{equation}
\mathcal{L}_{\mathrm{tree}}
\;=\;
\gamma\,\mathrm{CE}\!\left(\hat{x},x\right)
\;+\;
\eta\,\frac{1}{|M|}\sum_{m \in M}\Big( \|\hat{L}_m - L_m\|^2 + \|\hat{R}_m - R_m\|^2 \Big)
\;+\;
\beta\,\mathrm{KL}.
\label{eq:tree}
\end{equation}
Here $x$ is the input sentence and $\hat{x}$ its reconstruction; $M$ is the set of merge nodes; $L_m$ and
$R_m$ are the child latents at node $m$ and $\hat{L}_m, \hat{R}_m$ the unmerger's predictions; and $\mathrm{KL}$
is the KL divergence from $N(0,I)$ averaged over the leaf posteriors plus the same quantity averaged over the
merge posteriors. The reconstruction cross-entropy,
weighted by $\gamma$, is the main training signal for the encoder. The one-step term, weighted
by $\eta$ and averaged over merge nodes, asks each parent to be invertible into its own children; it is
computed on the upward pass as each parent is formed. The KL, weighted by $\beta$, is a price per unit of
information kept in a latent, so that only what is necessary is retained.

When we compute the one-step term we detach the gradients of the parent and of both target children, so that
the only network left in the gradient path is the unmerger. This term trains the unmerger and never reaches
the merger or the lifter. We chose this to stop the
encoder from preferring merges that are easy to reverse one step at a time rather than merges that compress
well; otherwise the encoder could lower the loss with merges that are locally invertible but do not follow
constituent structure, bypassing the bottleneck. So
the encoder is shaped only by the root reconstruction and the KL.

This objective needs a tree to be given, which the base models do not have. \Cref{sec:chart} removes that
requirement by scoring every tree instead of choosing one. We return to \Cref{eq:tree} only for the
distilled student, once the consensus supplies the trees (\Cref{sec:distill}).

\subsection{Inducing structure by marginalising over trees with the CKY inside chart}
\label{sec:chart}
We remove the need for a given tree with an inside chart. Instead of deciding which pair to merge, we score
every binary tree at once. Scoring each tree separately would take exponentially many passes, but trees share
sub-analyses, so a chart computes each span once and covers all trees in $O(n^3)$ time. Training raises the
score of trees whose merges can be inverted with low error. \Cref{fig:chartcell} shows one chart cell.

\subsubsection{The inside recursion}
The unit of the chart is a span, a contiguous stretch of the sentence, and each span carries a score and a
representation. At a single token the representation is the lifter's latent and the score is zero. For a
longer span, every split point is one way the span could have been built, and each split produces its own
candidate parent. We score a split much as we compute the one-step term, except that only the
parent is detached, so this score also trains the representations of the two child spans. We merge the two child spans, unmerge
the result, measure the squared error against the two children, and negate it. A split's total score adds the
two child-span scores to this local score,
\begin{equation}
s(i,j) \;=\; \operatorname*{logsumexp}_{i<k<j} \Big[\, s(i,k) + s(k,j) \;-\; \big(\|\hat{L}-r(i,k)\|^{2} +
\|\hat{R}-r(k,j)\|^{2}\big) \Big],
\label{eq:inside}
\end{equation}
where $r(i,j)$ is the representation of span $(i,j)$ and $\hat{L}, \hat{R}$ are
$\mathrm{Unmerge}(\mathrm{Merge}(r(i,k), r(k,j)))$, with base case $s(i,i+1)=0$. The span's score is the
log-sum-exp of its splits, which sums over all trees rooted at that span, so the root score $s(0,n)$ aggregates,
over all trees rather than one we picked, how accurately each merge can be inverted. Maximising it therefore
trains the model on every tree at once, each tree contributing to the gradient in proportion to its
exponentiated score.

\begin{figure}[t]
\centering
\begin{tikzpicture}[
  >=Latex,
  box/.style={draw, rounded corners, minimum height=6.5mm, inner sep=4pt, font=\small}
]
  \node[box] (lik) at (0,0) {$r(i,k)$};
  \node[box] (rkj) at (2.6,0) {$r(k,j)$};
  \node[box, fill=black!6] (merge) at (1.3,1.2) {Merge};
  \node[box] (par)   at (1.3,2.4) {parent $p_k$};
  \node[box, fill=black!6] (unm) at (1.3,3.6) {Unmerge};
  \node[box] (lhat)  at (0,4.8) {$\hat{L}$};
  \node[box] (rhat)  at (2.6,4.8) {$\hat{R}$};
  \draw[->] (lik) -- (merge);
  \draw[->] (rkj) -- (merge);
  \draw[->] (merge) -- (par);
  \draw[->] (par) -- (unm);
  \draw[->] (unm) -- (lhat);
  \draw[->] (unm) -- (rhat);
  \draw[dashed,<->] (lhat) to[bend right=45] (lik);
  \draw[dashed,<->] (rhat) to[bend left=45] (rkj);
  \node[font=\scriptsize, anchor=east] at (-1.1,2.4) {energy};
  \node[font=\scriptsize, anchor=west] at (3.7,2.4) {energy};
  \node[align=center, font=\small] at (1.3,-1.2)
    {$\mathrm{comb}_k = s(i,k)+s(k,j)-\big(\|\hat{L}-r(i,k)\|^2+\|\hat{R}-r(k,j)\|^2\big)$};
  \node[align=center, font=\small] at (1.3,-1.9)
    {$s(i,j)=\operatorname{logsumexp}_{k}\,\mathrm{comb}_k$};
\end{tikzpicture}
\caption{One chart cell for span $(i,j)$ at split $k$. The two child spans are merged into a candidate parent,
which the unmerger tries to invert; the negative squared error (``energy'') is the local score. The span score
is the log-sum-exp over splits, and the highest-scoring split is stored as a back-pointer for parsing.}
\label{fig:chartcell}
\end{figure}

\subsubsection{Straight-through aggregation}
Two different quantities are involved here, and only one of them is soft. The span \emph{score} is a soft
sum, the log-sum-exp over splits of \Cref{eq:inside}, which is what keeps the objective
differentiable. The \emph{representation} passed up from a span, however, must be a single vector, so for it
we select one split with a hard Gumbel sample on the forward pass while letting the gradient flow through the
soft weights. This is the straight-through estimator. The score is therefore soft while the chosen merge stays a
genuine discrete decision. The Gumbel temperature is annealed from 2.0 to 0.5 over 10{,}000 steps; since the
base models train for only 2{,}500 steps, the temperature only falls from 2.0 to about 1.63 during their training,
the high, exploratory part of that schedule. We accumulate only the selected split's KL, since the same
hard weights choose both the representation and its KL. Two alternatives are available, a plain argmax and a soft blend over the candidate splits. We compare all three and justify this choice in Experiment~1.

\subsubsection{Extracting a parse}
While filling each cell the chart also records which split scored highest. To read off the model's parse we
follow these back-pointers down from the root, so one mechanism gives both the training signal (a sum over
trees) and a single parse at test time (an argmax over trees); at test time we replace the Gumbel sample with a
plain argmax, which makes parsing deterministic. Throughout, the chart uses the latent means, so it is
deterministic, and $\beta$ pulls the means towards zero and the variances towards one without adding sampling
noise to the chart.

\subsubsection{Training objective for the base models}
\label{sec:baseobj}
The base models have no given tree, so we train them with a different objective that scores every tree with
the chart:
\begin{equation}
\mathcal{L}_{\mathrm{chart}}
\;=\;
-\,\lambda\, \mathbb{E}_{x}\big[\, s(0,n) \,\big]
\;+\;
\gamma\,\mathrm{CE}\!\left(\hat{x},x\right)
\;+\;
\beta\,\mathrm{KL}.
\label{eq:chart}
\end{equation}
It has three self-supervised terms. It maximises the root inside score (weight $\lambda$), reconstructs the leaves
after decoding the highest-scoring (argmax) tree from the root representation (weight $\gamma$), and applies the KL (weight $\beta$). Unlike
\Cref{eq:tree}, the one-step reconstruction is not a separate term here; it lives inside the chart
recursion as the local score of \Cref{eq:inside}. \Cref{eq:tree,eq:chart} are two
different objectives used at two stages. \Cref{eq:chart} trains the base models, which have no tree,
and \Cref{eq:tree} trains the student, once the trees are given.

A base model's held-out parse-validity often peaks early in training and can decline as training
continues. Rather than prevent this decline, we end training early. The learning rate is held constant for the
first 1{,}500 steps and then decayed along a cosine to near zero by step 2{,}500, so the parameters stop changing
at whatever solution has been reached. This is a practical measure, not an explanation of the decline, and \Cref{sec:lottery} reports the
instability that motivates it.

\subsection{From many parsers to one treebank}
\label{sec:select}
We train many base models as parsers, because even with the same training set their quality varies. Because we
want to keep the method unsupervised, we cannot score them all against gold trees and keep the best; instead we select
the best ones without supervision and then combine their parses. To select, we rank the models by their mean
agreement with every other model, where the agreement of two models is the fraction of training sentences on
which their parse trees are identical (peer agreement), and keep the third with the highest peer agreement. We use
this because we assume that bad models fail in mutually disagreeing ways while good models converge on similar,
correct parses, so agreement is a usable proxy for quality. To combine, each selected model votes for the
constituents in its parse of each sentence, and we build the silver tree from the bracketing with the most
total span support, where the support of a span is the number of selected models whose tree contains it, which is minimum Bayes risk decoding at the level of constituents \citep{goodman1996parsing}. We work at the level of
constituents rather than whole trees because, on ambiguous data, two good models may produce different but
equally valid trees; a whole-tree vote would split their support and could discard a good model, whereas a
constituent-level vote lets partially agreeing models reinforce the spans they share (\Cref{sec:combine}). Finally,
we use agreement only to select the models and then let them vote uniformly, since agreement already decides which models are
retained, and weighting by it as well would further favour the models closest to the majority, whether or not
the majority is correct.

\subsection{Distillation into a single model}
\label{sec:distill}
The consensus gives us a silver treebank, one tree for every training sentence. We train a fresh student model
on it with our fixed-tree objective (\Cref{eq:tree}), teacher-forced on the silver trees, that is, trained with each silver tree given as the merge order. We distil a
student because the base models are good parsers but poor reconstructors, a consequence of the tight bottleneck
that forces them to find structure. Fixing the structure first lets the student devote its capacity to
reconstruction rather than on discovering how to bracket a sentence. The two stages therefore act as a
curriculum. A tight bottleneck in the first stage induces the grammar at the cost of reconstruction, and a
looser bottleneck in the second stage learns to reconstruct on the now-fixed trees. Grammar discovery and
reconstruction are separated in time rather than traded off inside one model, using the same architecture
under different objectives at the two stages. Whether one model could instead do both on its own, by training
longer or loosening its bottleneck rather than separating the stages, is tested directly in Experiment~6. The student
is given a larger latent (embedding dimension 48), a
smaller KL weight ($\beta = 0.002$) and a longer schedule (12{,}000 steps) than the base models. The
reconstruction this recovers is reported in Experiment~4, and the sweep of the student's KL weight that places
this value on the reconstruction-favouring end of the trade-off in Experiment~5.

\section{Experiments}
\label{sec:experiments}

We report six experiments. The first fixes the way the chart aggregates a span. The second measures how a
single base model behaves and exposes the seed lottery. The third studies how the base models should be
combined. The fourth is the main result across seven independent pools. The fifth asks what the latent space
represents. The sixth asks whether a single model trained longer under different learning-rate and KL schedules could replace the two-stage
design. A shared setup precedes them, and a section on limitations and future work follows.

\subsection{Setup}
\label{sec:setup}
Our grammar is a small ambiguous probabilistic grammar in the style of English, with 19 terminal words in six
categories (subjects, verbs, gradable adverbs, non-gradable adverbs, grade modifiers and verb qualifiers)
and 7 non-terminals. Its ambiguity comes from recursive rules. A grade may modify a grade, and a verb may take
qualifiers and adverbs, so many surfaces have several distinct valid parses. 63.1\% of surfaces admit more
than one, and the average surface admits 4.34. Ambiguity grows sharply with length. Sentences of length two
and three are unambiguous, while at length eight 94.8\% of surfaces are ambiguous and the average admits 10.8
parses. Sentences are of length 2 to 8, sampled by
derivation, so each comes with the gold tree that generated it. A single surface can occur as several
sentences, one for each distinct derivation the grammar assigns it, so the number of sentences exceeds the
number of distinct surfaces. We split by surface, holding out 10\% of the distinct surfaces at each length;
this gives 1370 distinct surfaces in total, which yield 1510 training and 174 held-out sentences.

We evaluate the distilled model with three metrics. Parse-validity is the fraction of held-out sentences whose
predicted tree is a valid parse under the grammar (the baselines in \Cref{tab:baselines} are averaged over
surfaces instead). Recon-exact is the fraction whose tokens, decoded from the
single root vector along the tree with the lowest $\mathcal{L}_{\mathrm{tree}}$, are recovered exactly. Gold-match is the fraction whose predicted tree equals the
sentence's gold tree. Parse-validity is our primary metric, because our aim is to recover any valid analysis
rather than one particular tree; gold-match is deliberately stricter, since under ambiguity the model can pick
a different but valid parse, and it tells us how often it recovers the exact sampled tree. Gold trees are used
only in evaluation, never in training, selection, or the consensus.

We also report the standard unlabelled bracketing F1 in \Cref{sec:mainresult}, but treat it as secondary. Under an
ambiguous grammar F1 against a single gold tree is a lower bound rather than a measure of correctness, since a
model that returns a different but equally valid parse is penalised for the sampler's arbitrary choice.
Parse-validity and gold-match instead give upper and lower bounds on correctness (validity
credits any valid analysis, gold-match demands the exact sampled one), and because the grammar is small
enough to enumerate, both are computed exactly rather than estimated. We take recovering a valid constituent
structure as the target, which is what parse-validity measures.

A validity score is only meaningful against a baseline, so we measure three. Because sentences are short
enough to enumerate every binary tree exhaustively, the random-tree baseline can be computed exactly rather
than sampled; for each surface it is the number of valid parses divided by the number of binary trees over its
tokens. \Cref{tab:baselines} gives all three on the held-out set. Right-branching is the strongest (0.416, against 0.150 for a random tree),
which is expected because the top-level rule splits off a one-word subject and the qualifier and grade rules
recurse to the right, and it is the reference baseline for the rest of the
paper. The per-length breakdown matters more than the aggregate. Right-branching is perfect at
length three and falls to 0.080 by length eight, while a random tree falls from 0.500 to 0.027. Everything
we report later should be read against those columns, because the difficulty grows sharply with length.

\begin{table}[t]
\centering
\caption{Baselines on the held-out set (137 surfaces). ``Random'' is the exact expected validity of a uniformly
random binary tree, obtained by enumeration rather than sampling. Overall figures are averaged over surfaces.}
\label{tab:baselines}
\begin{tabular}{lcccccccc}
\toprule
& Overall & $L{=}3$ & $L{=}4$ & $L{=}5$ & $L{=}6$ & $L{=}7$ & $L{=}8$ \\
\midrule
Random binary tree & 0.150 & 0.500 & 0.256 & 0.137 & 0.094 & 0.046 & 0.027 \\
Left-branching     & 0.007 & 0.000 & 0.000 & 0.000 & 0.000 & 0.000 & 0.000 \\
Right-branching    & 0.416 & 1.000 & 0.800 & 0.360 & 0.320 & 0.240 & 0.080 \\
\bottomrule
\end{tabular}
\end{table}

We implement the whole pipeline in PyTorch and run it on the department's GPU cluster. A pool is one run of the full pipeline,
consisting of 24 base models trained from different random seeds and the student distilled from them. The hyperparameters are given in \Cref{tab:config}.

Three implementation constraints shaped the experiments. The chart is cubic in sentence length and is built
in Python for every training step, which is the dominant cost and the reason sentences are capped at length
eight; sentences are grouped by length within a batch so that each group can be charted as a single tensor
operation. Evaluation is more expensive than training, because scoring reconstruction requires enumerating
every binary tree of a sentence and running the model over each one, which is tractable only under the same
length cap. Finally, the chart runs on the latent means rather than on samples (\Cref{sec:chart}); only the
student, which trains on fixed trees, samples its latents.

\begin{table}[t]
\centering
\caption{Hyperparameters for the base models and the distilled student. Each pool trains 24 base models, of
which the 8 with the highest peer agreement are selected; results are reported over seven independent pools.
The KL weight is warmed up, so the base models, which stop at step 2500, end at $\beta \approx 0.032$.}
\label{tab:config}
\begin{tabular}{lcc}
\toprule
Hyperparameter & Base model & Student \\
\midrule
Embedding dimension            & 32                  & 48                 \\
Hidden dimension               & 128                 & 128                \\
Learning rate                  & $2\times10^{-3}$    & $2\times10^{-3}$   \\
Training steps                 & 2500                & 12000              \\
Batch size                     & 32                  & 32                 \\
KL weight $\beta$              & 0.04                & 0.002              \\
KL warm-up                     & \multicolumn{2}{c}{$0.2\beta$ to step 1000, then linear to $\beta$ by step 3000} \\
Root-score weight $\lambda$    & 0.2                 & --                 \\
Reconstruction weight $\gamma$ & 5.0                 & 5.0                \\
One-step weight $\eta$         & --                  & 1.0                \\
Gumbel temperature             & $2.0\!\to\!0.5$ over 10000 & --          \\
LR schedule                    & hold then decay     & constant           \\
\bottomrule
\end{tabular}
\end{table}

\subsection{Experiment 1, choosing the aggregation mode}
\textbf{Setup.} The chart can pass a representation up from a span in one of three ways, the straight-through
hard sample we use, a plain argmax, or a soft blend that averages the candidate splits under a softmax over
their scores. We ran the full pipeline once per mode on the same twenty-four seeds and the same data split, so
that only the aggregation rule varied. As a control, the straight-through run reproduces the base models of pool 76--99
exactly, confirming that the three runs were otherwise identical.\footnote{Maximum absolute difference in held-out parse-validity across all
twenty-four matched seeds was $0.0000$.}

\begin{table}[t]
\centering
\caption{Aggregation ablation. Twenty-four base models per mode on matched seeds and one data split, so
initial weights and batch order are identical within each row and only the aggregation rule differs. Base-model
figures are medians over the twenty-four models; consensus and student figures are for the pipeline built from
them.}
\label{tab:aggregation}
\resizebox{\textwidth}{!}{\begin{tabular}{lccccc}
\toprule
Aggregation & Base validity & Base recon. & Consensus & Student validity & Student recon. \\
\midrule
Straight-through & \textbf{0.552} & 0.147 & \textbf{0.985} & 0.989 & \textbf{0.943} \\
Hard (argmax)    & 0.328 & \textbf{0.233} & 0.981 & \textbf{0.994} & 0.925 \\
Soft (blend)     & 0.443 & 0.118 & 0.931 & 0.948 & 0.920 \\
\bottomrule
\end{tabular}}
\end{table}

\textbf{Results.} At the level of the individual base model, straight-through gives the highest median
parse-validity, 0.552 against 0.443 for the blend and 0.328 for the plain argmax. The blend is the worst
reconstructor at 0.118. At the level of the finished pipeline the three modes largely converge, and the
distilled students differ by less than one held-out sentence between straight-through and argmax.

\textbf{Discussion.} We keep straight-through for two reasons. It gives the best base models and the best final
reconstruction, and it is the only one of the three in which the merge is a hard decision and still receives a gradient,
since a plain argmax passes no gradient through the representation it selects, and a soft blend makes no hard
decision. Although averaging
incompatible splits cannot represent any single tree, the pipeline built on the soft blend still reaches a
student validity of 0.948. That the finished pipeline is robust to all three choices is useful in practice, because the
method does not depend on a delicate setting of the aggregation rule.

\FloatBarrier
\subsection{Experiment 2, a single model and the seed lottery}
\label{sec:lottery}
\textbf{Setup.} We trained the four seed-varying pools of Experiment~4, 24 base models each on a single data
split, with the same configuration throughout and only the random seed changing.

\begin{figure}[t]
\centering
\includegraphics[width=0.8\textwidth]{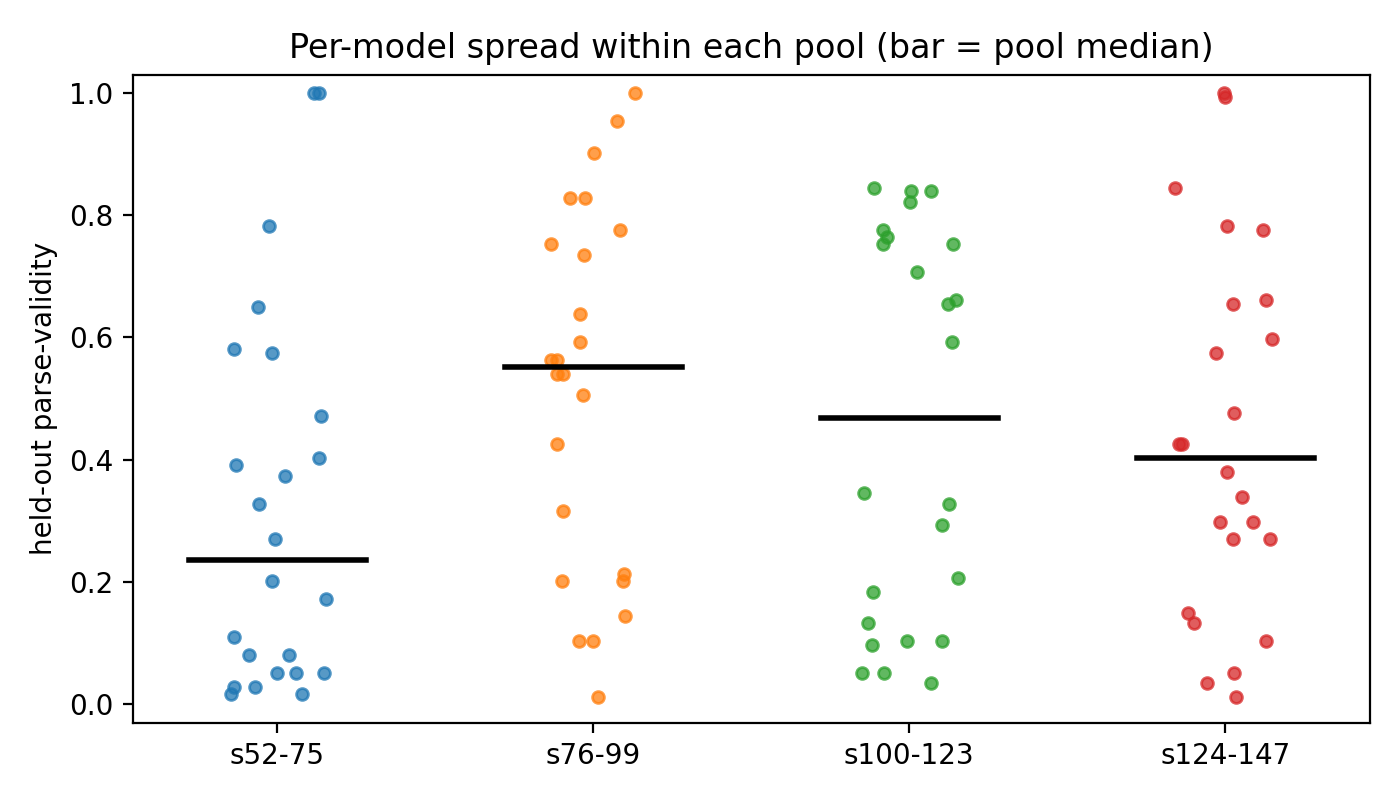}
\caption{The seed lottery. Held-out parse-validity of the 24 base models in each of the four seed-varying
pools (labelled by their seeds, so s52-75 is pool 52--75). The spread within a pool is wide even though the
models differ only in random seed.}
\label{fig:lottery}
\end{figure}

\textbf{Results.} The base models' held-out parse-validity ranges from 0.01 to 1.00, and pool medians sit
between 0.24 and 0.55, so the method can discover the grammar completely in its best runs while a single model
is not reliable. \Cref{fig:lottery} shows the per-model spread across the four pools.

\textbf{Discussion.} Sweeps of the KL weight, embedding dimension, batch size and learning-rate schedule during
development did not remove this volatility, which is why we combine 24 base models by selection, consensus and distillation. The spread is consistent with
the Recursive Bayesian Network view (\Cref{sec:rbn}), where marginalising over structures gives a non-convex
objective with one component per tree, so different seeds converge to different local optima and gradient descent is
not guaranteed to find the best one. By contrast, DIORA reports low seed variance; we hypothesise that our single-vector bottleneck is the
source of the instability, but we have not tested this.

\subsection{Experiment 3, combining models}
\label{sec:combine}
\subsubsection{Constituent-level consensus versus whole-tree voting}
\textbf{Setup.} On each of the seven pools we combined the eight selected models both ways, span-based minimum
Bayes risk, which the pipeline uses, and a whole-tree majority vote, and scored the resulting silver treebank
against the grammar on the training surfaces. We also distilled a student on each silver treebank in turn, with
training otherwise identical, to test whether the choice affects the distilled student.

\textbf{Results.} Span-MBR matches or beats the vote in all seven
pools, raising the mean consensus from 0.867 to 0.933, with a worst case of an exact tie rather than a
regression. The gain concentrates where the whole-tree vote is weakest. In pool 52--75 the vote returns only
0.616 even though the pool contains a model with parse-validity 1.000 on the training surfaces, and combining at the constituent level lifts that
pool to 0.887. The whole-tree vote never beats its own best member in any of the seven pools. The advantage
carries over to the distilled student, as the span-MBR student matches or beats the whole-tree-vote student on held-out
validity in all seven pools, by 0.068 on average and by 0.287 in pool 52--75.

\textbf{Discussion.} In pool 52--75 a whole-tree vote discards the trees of the best model, because the strong models
agree on the constituents but disagree over which equally valid whole tree to report, so their support is
split across different trees. A vote can only return a tree that some model already proposed, whereas span-MBR assembles
the best-supported constituents into a bracketing that need not equal any single input tree. On ambiguous data,
where good models often return different but equally valid trees, the combiner must therefore work on
constituents; otherwise the student inherits the lower-quality treebank. This is consistent with \citet{shayegh2024ensemble}, who combine
ensembles of different parsers in the same way.

\subsubsection{How many models, and which ones}
\begin{figure}[t]
\centering
\includegraphics[width=0.8\textwidth]{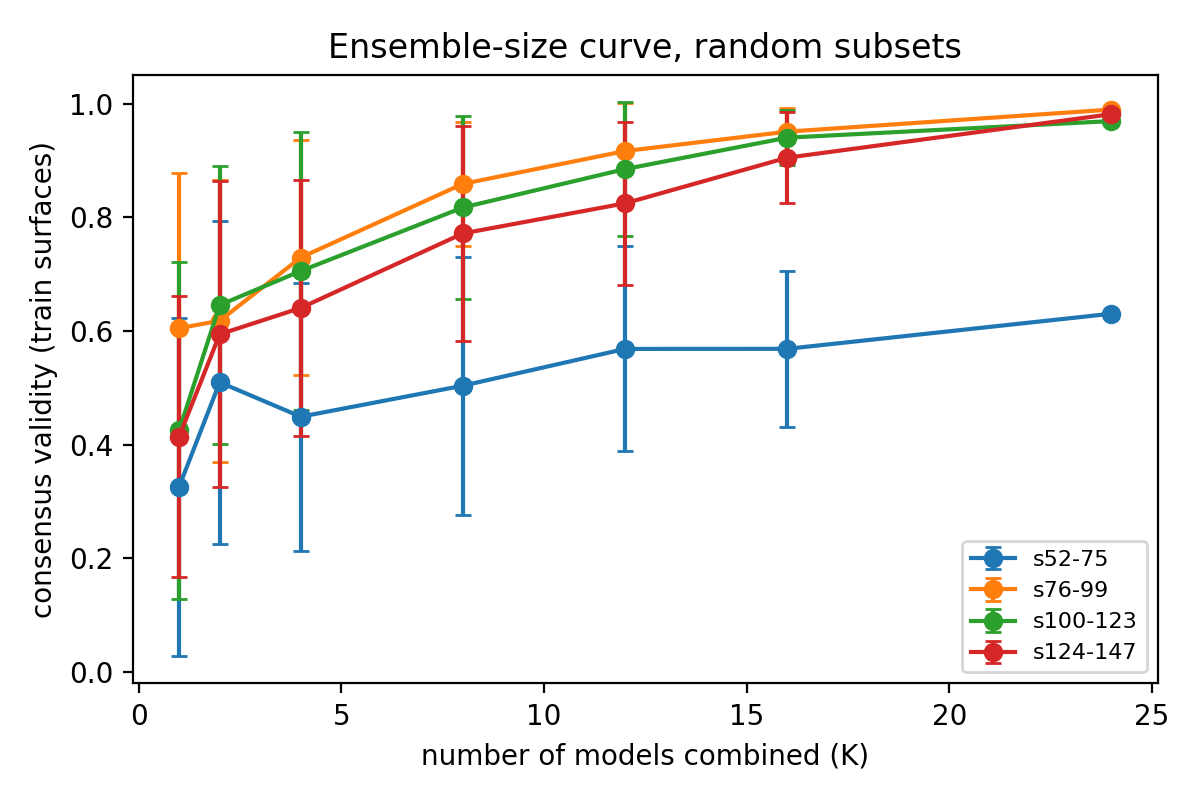}
\caption{Consensus validity against ensemble size when models are drawn at random, averaged over sixty draws
per point, for the four seed-varying pools. Every pool improves with size, apart from one dip in pool 52--75, and
the spread narrows. Compare \Cref{tab:ksweep}, where models are added in order of decreasing peer agreement and
the consensus does not rise with size.}
\label{fig:ensemble}
\end{figure}

\textbf{Setup.} Two questions remain, how much of the benefit comes from combining many models and how much
from combining the right ones. We separate them with two sweeps. The first, over the four seed-varying pools, adds
models drawn at random, averaging sixty random subsets of each size (\Cref{fig:ensemble}). The second, over all
seven pools, adds them in the order the selection rule picks, ranked by peer agreement (\Cref{tab:ksweep}).

\textbf{Results.} Combining more models helps when they are drawn at random. Every pool improves with size, apart from
one dip in pool 52--75 between two and four models, and the spread narrows sharply; in pool 76--99 the mean rises from 0.605 with a single model to 0.989
with all twenty-four, while the standard deviation across draws falls from 0.273 to zero. Combining the models
the selection rule actually picks behaves differently. The curve is not monotone, and its shape differs
between pools; pool 52--75 falls from 0.999 to 0.630 as models are added, while pool 100--123 rises from
0.863 to 0.969. Peer agreement, the signal the selection rule ranks by, tracks gold-scored parse-validity closely. Across the
ninety-six models in the first four pools their rank correlation is 0.863 (Spearman), and it never falls below
0.838 in any single pool.

\begin{table}[t]
\centering
\caption{Consensus validity against the number of models retained, ranked by peer agreement, over all seven
pools. Unlike \Cref{fig:ensemble}, which draws models at random, this adds them in decreasing order of peer agreement. $K{=}8$ is
the value the pipeline uses.}
\label{tab:ksweep}
{\small
\begin{tabular}{lccccccccc}
\toprule
Models retained & 2 & 4 & 6 & \textbf{8} & 10 & 12 & 16 & 20 & 24 \\
\midrule
Mean over pools  & 0.951 & 0.903 & 0.928 & \textbf{0.933} & 0.931 & 0.920 & 0.884 & 0.899 & 0.887 \\
Worst pool       & 0.796 & 0.796 & 0.819 & \textbf{0.844} & 0.797 & 0.716 & 0.663 & 0.682 & 0.630 \\
\bottomrule
\end{tabular}}
\end{table}

\textbf{Discussion.} The random-draw sweep shows the variance reduction expected of an ensemble. The two sweeps diverge because they do different things. Adding a random model adds diversity, whereas
adding a peer-ranked model adds a model the ranking has already judged worse than everything above it, so once
the highest-ranked models are included, each further model tends to lower the consensus. That peer agreement predicts gold-scored validity so well means an
unsupervised quantity we can compute is a reliable stand-in for the supervised one we cannot. We therefore
retain eight models. Eight is not the value with the best average, since retaining two is marginally better on
the mean, but it is the value with the best worst case, and we prioritise the worst pool over the mean because the purpose of the
pipeline is a reliable result from any pool. Retaining two gives up 0.048 in the worst pool to gain 0.018 on average. The choice is also
insensitive; across the range from six to twelve models the average moves by only 0.012, so eight sits on a
plateau rather than at a spike, which matters because the value was chosen with knowledge of these gold-scored
results rather than on a separate development set (\Cref{sec:limitations}).\footnote{The figure for two models should not be read as evidence that a
two-model ensemble combines well. With two models the combiner almost always returns the better of the two
rather than anything new. Across the seven pools its mean advantage over the better member is $+0.0008$, and
in one pool the consensus reproduces the better model's tree on all 549 sentences where the two disagree. That
column measures how good a model the selection rule can find, not what combining achieves.}

\subsection{Experiment 4, the main result across seven independent pools}
\label{sec:mainresult}
\textbf{Setup.} We ran seven independent pools of 24 base models each (\Cref{tab:sevenpool}). Four vary only the random
seed on a single data split and so measure initialisation and optimisation variance. The other three vary the
data split as well. Our sampler is reseeded for each, so these are fresh draws of sentences from the same
grammar rather than re-partitions of the same corpus, and the four distinct splits give held-out sets of 174,
171, 165 and 172 sentences. Together they separate the two sources of variability the result might have
been sensitive to.

\textbf{Results.} Student results are similar under both sources of variation. Distilled students reach 0.839 to 0.989 across seeds and 0.855 to 0.994 across
splits, and the two groups have almost the same mean. The pipeline reduces the 0.01 to 1.00 spread of single
base models to 0.839 to 0.994, and redrawing the data does not widen it.

The students also reconstruct far better than the base models they are distilled from. Across the four seed-varying pools the
median base model recovers its input exactly on only 0.147 to 0.198 of held-out sentences, and the worst on
0.023, whereas the students reach 0.925 to 0.971. This roughly fivefold gain comes from fixing the structure
before learning to reconstruct, which is the point of training a second model with a looser bottleneck (Experiment~6 tests a
single-model alternative).

Selection does not improve every pool; its benefit is concentrated in the weakest pools. Three pools are
already strong without it, and on those it lowers the result, by up to 0.10 (pool 100--123 falls from 0.969 to
0.867). It raises the weakest pools instead. The worst pool, 52--75, improves from 0.630 to 0.887 and split 1 from 0.805 to 1.000, and across all seven the mean rises
from 0.886 to 0.933 while the minimum rises from 0.630 to 0.844. Peer-agreement selection therefore raises the worst case
rather than improving every pool, at a cost on pools that are already strong.

One limitation bounds what the ensemble can claim. The consensus never
beats the best individual model in its pool, unlike the ensembles of different parsers in \citet{shayegh2024ensemble}. Pools 52--75 and 124--147 each contained a model with parse-validity
1.000 on the training surfaces, and the consensus matched neither. What the consensus does is reliably approach the validity of a good model
without knowing which model that is, and that is the appropriate comparison. Against the best model
identifiable without gold trees (the one our own selection rule ranks first), consensus wins over the four
seed-varying pools on training surfaces, 0.92 against 0.80 on average and by much more in the worst case,
0.867 against 0.555. The consensus therefore acts as a
variance-reduction step rather than a source of gains beyond the best member. Because parse quality cannot be
measured without gold trees, the best model in a pool cannot be identified at training time; the ensemble
recovers near-best performance reliably from a pool whose individual outcomes are, by the seed lottery,
unpredictable.

A final check tests whether the distilled model has learned the grammar or merely memorised the silver trees
it was trained on. The split is by surface, so no held-out sentence appears in training, but the student could
in principle still overfit the training brackets. The evidence is against memorisation. On the two pools
examined this way, 52--75 and 76--99, parse-validity on training surfaces (0.900 and 0.980) matches held-out validity (0.868 and
0.989) to within sampling noise, so there is no generalisation gap. The student reproduces the silver trees it
was given on 0.905 and 0.995 of training surfaces, so it is a faithful distillation whose grammar transfers to
unseen surfaces rather than a memorisation of the training trees.

\textbf{Discussion.} Across all seven pools, gold-match averages 0.494 while validity averages 0.927. This gap is
expected under ambiguity rather than a shortfall in the grammar the model has learned. 63.1\% of surfaces admit
more than one valid parse, and at length eight the average surface has about eleven of them, so a model that has
learned the grammar will frequently return a valid tree other than the one our sampler happened to draw.
Moreover, an ambiguous surface appears once per sampled derivation, each with its own gold tree, so a single
predicted tree can match at most one of them. The same gap appears in unlabelled bracketing F1, the standard parsing metric, computed across all
seven pools. F1 against the sampled gold tree averages 0.78 (range 0.74 to 0.81), but against the best-matching
valid parse it averages 0.98 (range 0.95 to 1.00). The model recovers almost every constituent of some valid
analysis, and the shortfall against gold is a choice among valid parses, not an error (these figures are for
our enumerable grammar and are not comparable to F1 on natural corpora). \Cref{tab:perlength} breaks the
students down by length and shows where the remaining errors sit. In two of the three pools validity is at least
0.92 at every length and 1.000 up to length five, and all three degrade at seven and eight,
exactly where the number of competing analyses grows fastest, from 3.5 per surface at length six to 5.8 at
seven and 10.8 at eight.

\begin{table}[t]
\centering
\caption{Seven independent 24-model pools. The first four vary only the random seed on one data split; the
last three redraw the data as well. The two consensus columns are the parse agreed over all 24 models and over
the eight retained by peer agreement; both are measured on training surfaces. Student columns are
held-out. Gold parses are used only to compute these numbers, never in training or selection.}
\label{tab:sevenpool}
\vspace{2pt}
\resizebox{\textwidth}{!}{\begin{tabular}{llccccc}
\toprule
& Pool & Consensus (all 24) & Consensus (top 8) & Student validity & Student recon. & Student gold-match \\
\midrule
\multirow{5}{*}{\rotatebox{90}{seed}}
& 52--75    & 0.630 & 0.887 & 0.868 & 0.971 & 0.483 \\
& 76--99    & 0.990 & 0.985 & 0.989 & 0.943 & 0.517 \\
& 100--123  & 0.969 & 0.867 & 0.839 & 0.925 & 0.443 \\
& 124--147  & 0.981 & 0.953 & 0.960 & 0.937 & 0.512 \\
& \emph{mean} & \emph{0.893} & \emph{0.923} & \emph{0.914} & \emph{0.944} & \emph{0.489} \\
\midrule
\multirow{4}{*}{\rotatebox{90}{split}}
& split 1   & 0.805 & 1.000 & 0.988 & 0.889 & 0.532 \\
& split 2   & 0.964 & 0.994 & 0.994 & 0.891 & 0.521 \\
& split 3   & 0.866 & 0.843 & 0.855 & 0.936 & 0.448 \\
& \emph{mean} & \emph{0.878} & \emph{0.946} & \emph{0.946} & \emph{0.905} & \emph{0.500} \\
\midrule
& \textbf{all seven} & \textbf{0.886} & \textbf{0.933} & \textbf{0.927} & \textbf{0.927} & \textbf{0.494} \\
& worst pool & 0.630 & 0.844 & 0.839 & 0.889 & 0.443 \\
\bottomrule
\end{tabular}}
\end{table}

\begin{table}[t]
\centering
\caption{Distilled students by sentence length, held-out, for the three seed-varying pools that were measured
this way. Both metrics degrade at lengths seven and eight. The right-branching baseline over
the same lengths is given for comparison.}
\label{tab:perlength}
\begin{tabular}{lcccccc}
\toprule
& $L{=}3$ & $L{=}4$ & $L{=}5$ & $L{=}6$ & $L{=}7$ & $L{=}8$ \\
\midrule
Validity, pool 76--99   & 1.000 & 1.000 & 1.000 & 1.000 & 0.960 & 0.960 \\
Validity, pool 124--147 & 1.000 & 1.000 & 1.000 & 0.929 & 0.920 & 0.920 \\
Validity, pool 100--123 & 1.000 & 0.936 & 0.846 & 0.833 & 0.720 & 0.760 \\
Reconstruction, pool 76--99 & 1.000 & 1.000 & 1.000 & 1.000 & 0.880 & 0.720 \\
\midrule
Right-branching baseline & 1.000 & 0.800 & 0.360 & 0.320 & 0.240 & 0.080 \\
\bottomrule
\end{tabular}
\end{table}

\subsection{Experiment 5, what the latent space represents}

\begin{figure}[t]
\centering
\includegraphics[width=\textwidth]{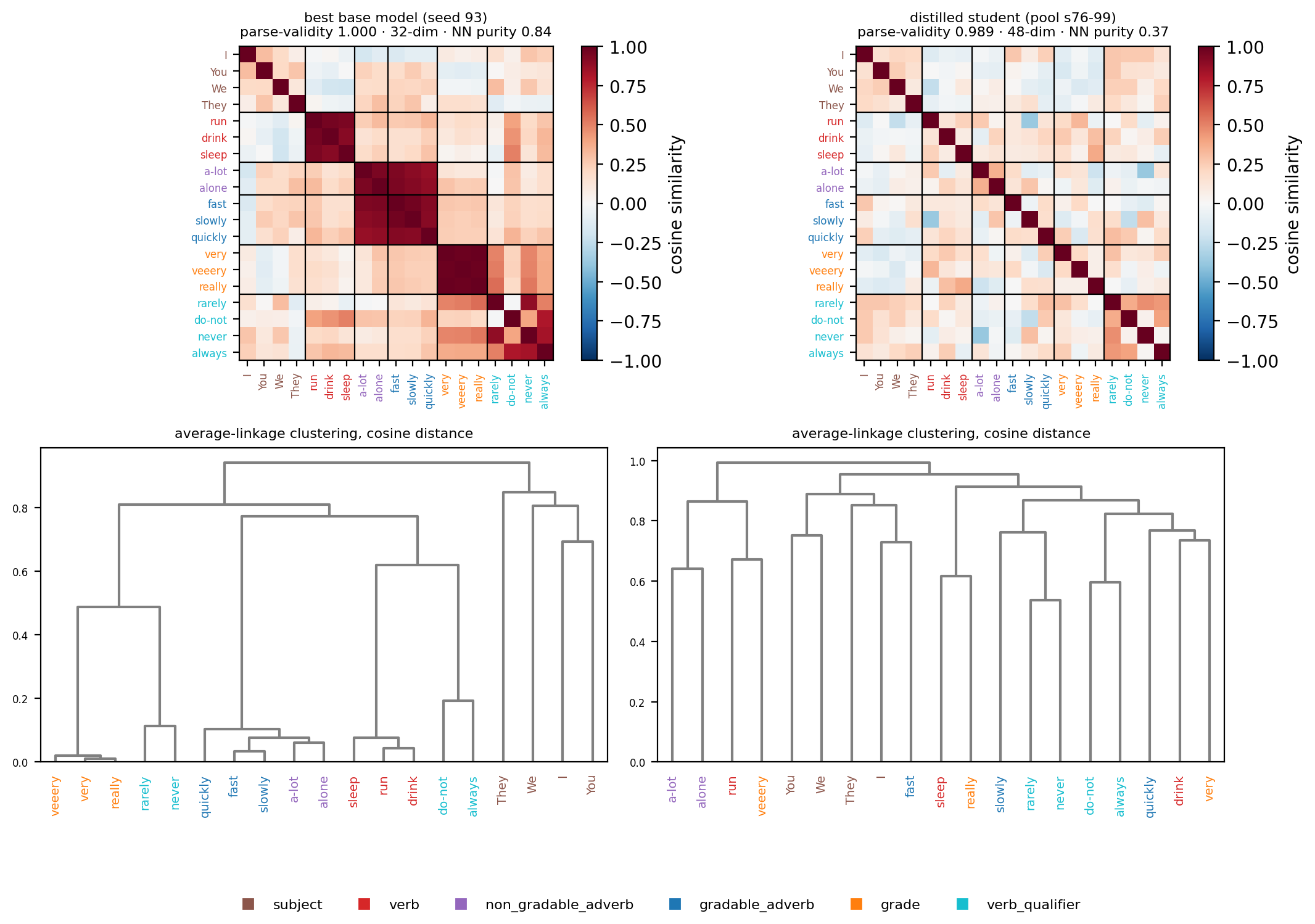}
\caption{Cosine similarity between the nineteen word embeddings, ordered by category, for the best base
model (held-out parse-validity 1.000) and the distilled student (0.989). The base model groups words by
category; the student does not, yet both parse well. A sweep of the student's KL weight (holding latent
size fixed) does not reproduce the clustering, so it is a by-product of how the base models induce
structure, not of the bottleneck and not a prerequisite for parsing.}
\label{fig:latent}
\end{figure}

\textbf{Setup.} The best base model and the distilled student learn the grammar equally well, their held-out
parse-validity being 1.000 and 0.989 respectively, but they represent the vocabulary differently. We compare
them through the cosine similarity between the nineteen word embeddings of each, ordered by category
(\Cref{fig:latent}).

\textbf{Results.} The best base model groups words tightly by syntactic type. Sixteen of the nineteen words
have a same-category nearest neighbour, and verbs, grade modifiers and gradable adverbs each form a distinct
block. The student shows little such grouping. Only seven of its nineteen words have a same-category nearest
neighbour, and for verbs, grade modifiers and gradable adverbs, none of the three words in each category has a
same-category nearest neighbour; its word vectors have pairwise cosine similarities close to zero.

\textbf{Discussion.} That a model can parse as well as the base model with little grouping of words by category tells us the
student's grammar is not carried by where its word vectors sit. The structure is in which vectors are
merged, not in their arrangement in the space. The natural explanation for the difference would be the
bottleneck, since the base models compress far harder than the student, but a controlled sweep argues
against it. Holding the latent size fixed and raising the student's KL weight from 0.002 to 0.05, past the base
models' 0.04, reconstruction degrades as expected, from 0.95 to 0.86, while category cohesion (the fraction of words whose nearest neighbour is in the same category) does not
increase, and at every setting the student groups words far less than the base model does. The base
models' word-category structure is therefore a property of how they are trained (they must induce the
grammar by scoring every tree, where the student receives its trees from the consensus) rather than of compression
pressure, though we do not isolate the cause further. The
network that performs the induction, with no category labels and no gold trees, organises the vocabulary by
syntactic type, which is consistent with chart-marginalised training recovering the grammar's categories and
not only its brackets. Whatever its cause, word-category geometry is a by-product of structure induction, not a prerequisite for parsing, and the
student parses without it.

\subsection{Experiment 6, a single model trained longer does not replace the two stages}
\textbf{Setup.} The pipeline separates structure induction from reconstruction across two models, a tight base
model and a looser student trained on fixed trees. A natural question is whether one model
could do both if it were trained for longer, removing the need for a second stage. We tested this on
four seeds that individually induce the grammar, whose base models reach held-out parse-validity between 0.90
and 1.00 and whose reconstruction is low as the tight bottleneck requires, near 0.13. For each seed we ran three
15{,}000-step strategies, all of which share the base model's tight first 2500 steps and then differ in how
training continues. The first restarts the learning rate at its maximum and keeps the bottleneck tight. The
second uses the same restart but drops the KL weight to the student's 0.002. The third uses no restart at all,
a single smooth learning-rate decay with the bottleneck held tight, which is equivalent to training the base model
for longer. The frozen base model is the reference. The three questions are whether more training recovers
reconstruction, whether the restart is needed, and whether loosening the bottleneck helps.

\begin{table}[!tb]
\centering
\caption{Four seeds that induce the grammar, each trained for 15{,}000 steps under three strategies that share
the tight first 2500 steps and then differ in how training continues. Restart resets the learning rate to its
maximum, continuous uses one smooth decay with no restart, and the loose arm drops the KL weight to 0.002. Base
is the frozen base model. Reconstruction recovers with training but plateaus below the student's 0.93, and no
strategy keeps the grammar while reaching 0.93 reconstruction. Validity and reconstruction are held-out.}
\label{tab:curriculum}
\begin{tabular}{lcccccccc}
\toprule
& \multicolumn{2}{c}{Base} & \multicolumn{2}{c}{Restart, tight} & \multicolumn{2}{c}{Restart, loose}
& \multicolumn{2}{c}{Continuous} \\
\cmidrule(lr){2-3} \cmidrule(lr){4-5} \cmidrule(lr){6-7} \cmidrule(lr){8-9}
Seed & Val. & Rec. & Val. & Rec. & Val. & Rec. & Val. & Rec. \\
\midrule
54  & 0.897 & 0.092 & 0.862 & 0.649 & 0.851 & 0.649 & 0.868 & 0.724 \\
74  & 0.994 & 0.172 & 0.966 & 0.644 & 0.805 & 0.621 & 0.948 & 0.615 \\
93  & 1.000 & 0.057 & 0.328 & 0.851 & 0.328 & 0.713 & 0.799 & 0.718 \\
134 & 1.000 & 0.167 & 1.000 & 0.713 & 0.989 & 0.615 & 0.994 & 0.638 \\
\bottomrule
\end{tabular}
\end{table}

\textbf{Results.} More training recovers reconstruction in every strategy (\Cref{tab:curriculum}), from
about 0.13 at the base model to between 0.61 and 0.85, but it plateaus there by around 12{,}000 steps and never
reaches the distilled student's
0.93. Loosening the bottleneck does not help. The loose arm is no better than the tight one on either metric, a
median change of $-0.011$ in validity and $-0.060$ in reconstruction. The restart is not needed and can be
harmful. The no-restart arm matches the restart on grammar across the four seeds, a median difference of
$0.000$ in validity, and it rescues seed 93, whose validity the learning-rate restart reduced from 1.000 to 0.328;
without the restart its validity is 0.799, at a reconstruction cost of 0.13 (0.718 against 0.851). Seed 93 also shows
the trade-off inside a single model most clearly. When its grammar collapses under the restart, to validity
0.328, reconstruction rises to the run's highest value, 0.851, whereas the no-restart run that keeps the
grammar at 0.799 reconstructs less, at 0.718.

\textbf{Discussion.} A single model reconstructs worse than the student not because its bottleneck is too
tight, since loosening it did not help, but because it must still induce structure while it reconstructs, and
the two objectives share its capacity. Seed 93 makes this visible, with reconstruction highest exactly where
validity has fallen. The student avoids the trade-off because it is trained on fixed trees and devotes all
of its capacity to reconstruction, which is why it reaches 0.93. Training one model for longer removes neither
the reconstruction plateau (0.61 to 0.85) nor the seed lottery, since these runs begin from seeds that already induce the grammar and still
spread, seed 93 ending between 0.33 and 0.80 depending only on the learning-rate schedule. The pipeline
provides the two properties the single model lacks, reliability from combining many seeded models and reconstruction
quality from a student trained on fixed trees. This is the two-stage separation of structure induction from
reconstruction, a frozen parser supplying the trees and a fresh model learning to reconstruct on them, and the
experiment shows that it is the separation, not a schedule inside one model, that lets the final model reach
both. We tested four seeds that induce the grammar and three schedules, so the claim is about this ceiling and
this lottery rather than a proof that no single-model schedule could ever do better.

\FloatBarrier
\subsection{Limitations and future work}
\label{sec:limitations}
Several factors limit what these results support; we quantify two of them.

The first concerns variance across pools. Our results are a distribution over pools rather than a
point estimate, and our original four pools shared a single data split, so they measured only initialisation
variance. The three additional pools in \Cref{tab:sevenpool} redraw the data, and the distilled students
span 0.855 to 0.994 across splits against 0.839 to 0.989 across seeds. Data variance is therefore no larger
than seed variance. Seven pools remains a small sample, and all seven come from one grammar.

The second is selection. The pipeline is unsupervised in the sense that no gold tree influences any model it
produces, but the rule for how many models to retain was chosen while we could see gold-scored results, which
is a form of tuning on the test set. We tried to remove that dependence by calibrating the rule on a separate,
unambiguous development grammar and transferring it unchanged, and it did not transfer. Over the four seed-varying
pools, the transferred rule averages 0.876 against 0.923 for the rule we use, and 0.796 against 0.867 in the worst case, because on an
unambiguous grammar validity means recovering the single correct tree, one of 429 at length eight, whereas on
our ambiguous grammar it means recovering any of roughly eleven, so the two grammars do not ask the same
question. We therefore quantify the dependence rather than leave it implicit. The selection rule is chosen with
knowledge of these pools, and closing that gap by fixing the rule without gold-scored feedback would cost
approximately 0.05 on average and 0.07 in the worst case. Every model the pipeline produces is still
unsupervised; the residual dependence is confined to this one scalar.

The remaining limitations are not quantified. Our metrics are not produced by identical procedures, since base
models are scored using the chart's own parse whereas the student's parse comes from enumerating every tree and
taking the one with the lowest $\mathcal{L}_{\mathrm{tree}}$ (\Cref{eq:tree}). The scale is small (19 words, sentences of length 2 to 8, a held-out set of 174
sentences), and exhaustive enumeration is tractable only under that bound, which is also what makes the
evaluation possible in the form we use. Validity is not correctness \citep{williams2018latent}. A tree can be
grammatical without being the analysis a linguist would choose, and the gold-match figure of 0.494 should be
read alongside the validity figure of 0.927. Because the student is distilled from the consensus, it can be
no better than the silver trees that supervise it, so any consistent but wrong bracketing the consensus settles
on is inherited by the student, and parse-validity, which accepts any valid tree, cannot detect it.

Finally, and most importantly, every result here is on a synthetic
grammar, and we make no claim about natural text. A grammar of nineteen words with six categories and a bounded
sentence length is a controlled setting chosen so that exhaustive enumeration, exact ambiguity counts and
exact random-tree baselines are all available. Those same properties are what natural corpora lack, which
is precisely why we begin here. The scope of the present contribution is that setting. We establish that
compression alone induces grammar, that the induction is a seed lottery, and that ensembling and distillation
convert the lottery into a single reliable parser. Extending this to natural text is the next step, and
\Cref{sec:conclusion} sets out the concrete obstacles (the loss of exhaustive enumeration and exact baselines chief among
them) that such an extension must address.

\section{Conclusion}
\label{sec:conclusion}
We asked whether a token-merging autoencoder can induce grammatical structure from raw tokens with no
supervision, by scoring all trees rather than choosing a merge, and whether that induction, if unstable, can be
stabilised without supervision. The answer to the first question is positive. The architecture recovers the grammar, in the best
runs with held-out parse-validity of 1.00, from a compression objective alone. The answer to the second is positive in aggregate. A
single model is a seed lottery, so we train many, keep the ones whose parses agree most often with the others', combine their parses span by span,
and distil the result into one model. That model reaches a mean of 0.93 parse-validity and 0.93 reconstruction on
held-out sentences, against a parse-validity of 0.42 for the strongest trivial baseline (right-branching), and it holds up across seven pools that
vary both the random seed and the data itself, with no gold trees used anywhere but evaluation.

Three further findings follow. First, unsupervised selection by peer agreement raises the worst case
rather than improving every pool. It raised our worst pool, from 0.63 to 0.89, and cost a strong one 0.10, raising the
across-pool mean and the worst case together. Second, the gain comes from which models are selected
rather than from how many are combined. Agreement among models predicts their gold-scored accuracy well, so adding
models the selector has already ranked low makes the consensus worse, even though adding models at random
makes it better. Third, the consensus never exceeds the best model in its pool; its benefit is a higher worst case, not gains
beyond the best member. It approaches the validity of a good model without knowing which model that is, and
against the best model we can identify without gold trees it wins in the worst case by 0.31 (0.867 against
0.555).

These results come from a small synthetic grammar and rest on a configuration chosen partly with knowledge of
results. \Cref{sec:limitations} sets out these limitations in full, including our unsuccessful attempt to remove the
dependence of the retention rule on gold-scored results.

Several directions follow. The selection rule still needs a way of being fixed without reference to the
evaluation data. Our attempt to calibrate it on a development grammar failed because validity means something
different on an unambiguous grammar than on an ambiguous one, so a transferable criterion would have to be
defined on a quantity that does not change meaning with the grammar. The next step is natural text,
which we have not attempted here and which will need an approximate chart, since exhaustive enumeration is
tractable only at short lengths; the same bound is what makes our evaluation exact, so moving to natural text
sacrifices exact evaluation. A more direct approach is a syntax-aligned objective (a context or cloze term in the manner of the
inside-outside autoencoder, or a neural grammar parameterisation), which would address the seed variance of
single models directly.
Our ensemble exists because single models are unreliable, and the results above suggest the ensemble can only
ever be as good as the best model it contains. Making that model reliable would be worth more than any
improvement to how they are combined.


\bibliography{refs}

@inproceedings{klein2004ccm,
  author    = {Klein, Dan and Manning, Christopher D.},
  title     = {Corpus-Based Induction of Syntactic Structure: Models of Dependency and Constituency},
  booktitle = {Proceedings of the 42nd Annual Meeting of the Association for Computational Linguistics (ACL)},
  year      = {2004},
  pages     = {478--485}
}

@article{baker1979inside,
  author  = {Baker, James K.},
  title   = {Trainable Grammars for Speech Recognition},
  journal = {The Journal of the Acoustical Society of America},
  volume  = {65},
  number  = {S1},
  year    = {1979},
  pages   = {S132}
}

@inproceedings{kim2019compound,
  author    = {Kim, Yoon and Dyer, Chris and Rush, Alexander M.},
  title     = {Compound Probabilistic Context-Free Grammars for Grammar Induction},
  booktitle = {Proceedings of the 57th Annual Meeting of the Association for Computational Linguistics (ACL)},
  year      = {2019},
  pages     = {2369--2385}
}

@inproceedings{drozdov2019diora,
  author    = {Drozdov, Andrew and Verga, Patrick and Yadav, Mohit and Iyyer, Mohit and McCallum, Andrew},
  title     = {Unsupervised Latent Tree Induction with Deep Inside-Outside Recursive Auto-Encoders},
  booktitle = {Proceedings of NAACL-HLT},
  year      = {2019},
  pages     = {1129--1141}
}

@inproceedings{shen2019ordered,
  author    = {Shen, Yikang and Tan, Shawn and Sordoni, Alessandro and Courville, Aaron},
  title     = {Ordered Neurons: Integrating Tree Structures into Recurrent Neural Networks},
  booktitle = {International Conference on Learning Representations (ICLR)},
  year      = {2019}
}

@inproceedings{shen2018prpn,
  author    = {Shen, Yikang and Lin, Zhouhan and Huang, Chin-Wei and Courville, Aaron},
  title     = {Neural Language Modeling by Jointly Learning Syntax and Lexicon},
  booktitle = {International Conference on Learning Representations (ICLR)},
  year      = {2018}
}

@inproceedings{choi2018gumbeltree,
  author    = {Choi, Jihun and Yoo, Kang Min and Lee, Sang-goo},
  title     = {Learning to Compose Task-Specific Tree Structures},
  booktitle = {Proceedings of the AAAI Conference on Artificial Intelligence},
  year      = {2018},
  pages     = {5094--5101}
}

@article{williams2018latent,
  author    = {Williams, Adina and Drozdov, Andrew and Bowman, Samuel R.},
  title     = {Do Latent Tree Learning Models Identify Meaningful Structure in Sentences?},
  journal   = {Transactions of the Association for Computational Linguistics (TACL)},
  volume    = {6},
  year      = {2018},
  pages     = {253--267}
}

@inproceedings{socher2011rae,
  author    = {Socher, Richard and Pennington, Jeffrey and Huang, Eric H. and Ng, Andrew Y. and Manning, Christopher D.},
  title     = {Semi-Supervised Recursive Autoencoders for Predicting Sentiment Distributions},
  booktitle = {Proceedings of the Conference on Empirical Methods in Natural Language Processing (EMNLP)},
  year      = {2011},
  pages     = {151--161}
}

@inproceedings{socher2011parsing,
  author    = {Socher, Richard and Lin, Cliff C. and Ng, Andrew Y. and Manning, Christopher D.},
  title     = {Parsing Natural Scenes and Natural Language with Recursive Neural Networks},
  booktitle = {Proceedings of the 28th International Conference on Machine Learning (ICML)},
  year      = {2011},
  pages     = {129--136}
}

@inproceedings{kingma2014vae,
  author    = {Kingma, Diederik P. and Welling, Max},
  title     = {Auto-Encoding Variational {Bayes}},
  booktitle = {International Conference on Learning Representations (ICLR)},
  year      = {2014}
}

@inproceedings{higgins2017beta,
  author    = {Higgins, Irina and Matthey, Loic and Pal, Arka and Burgess, Christopher and Glorot, Xavier and Botvinick, Matthew and Mohamed, Shakir and Lerchner, Alexander},
  title     = {{$\beta$-VAE}: Learning Basic Visual Concepts with a Constrained Variational Framework},
  booktitle = {International Conference on Learning Representations (ICLR)},
  year      = {2017}
}

@inproceedings{jang2017gumbel,
  author    = {Jang, Eric and Gu, Shixiang and Poole, Ben},
  title     = {Categorical Reparameterization with {Gumbel-Softmax}},
  booktitle = {International Conference on Learning Representations (ICLR)},
  year      = {2017}
}

@inproceedings{maddison2017concrete,
  author    = {Maddison, Chris J. and Mnih, Andriy and Teh, Yee Whye},
  title     = {The {Concrete} Distribution: A Continuous Relaxation of Discrete Random Variables},
  booktitle = {International Conference on Learning Representations (ICLR)},
  year      = {2017}
}

@article{bengio2013estimating,
  author  = {Bengio, Yoshua and L{\'e}onard, Nicholas and Courville, Aaron},
  title   = {Estimating or Propagating Gradients Through Stochastic Neurons for Conditional Computation},
  journal = {arXiv preprint arXiv:1308.3432},
  year    = {2013}
}

@article{hinton2015distilling,
  author  = {Hinton, Geoffrey and Vinyals, Oriol and Dean, Jeff},
  title   = {Distilling the Knowledge in a Neural Network},
  journal = {arXiv preprint arXiv:1503.02531},
  year    = {2015}
}

@inproceedings{goodman1996parsing,
  author    = {Goodman, Joshua},
  title     = {Parsing Algorithms and Metrics},
  booktitle = {Proceedings of the 34th Annual Meeting of the Association for Computational Linguistics (ACL)},
  year      = {1996},
  pages     = {177--183}
}

@inproceedings{shayegh2024ensemble,
  author    = {Shayegh, Behzad and Cao, Yanshuai and Zhu, Xiaodan and Cheung, Jackie C. K. and Mou, Lili},
  title     = {Ensemble Distillation for Unsupervised Constituency Parsing},
  booktitle = {International Conference on Learning Representations (ICLR)},
  year      = {2024}
}

@inproceedings{havrylov2019cooperative,
  author    = {Havrylov, Serhii and Kruszewski, Germ{\'a}n and Joulin, Armand},
  title     = {Cooperative Learning of Disjoint Syntax and Semantics},
  booktitle = {Proceedings of NAACL-HLT},
  year      = {2019},
  pages     = {1118--1128}
}

@inproceedings{yogatama2017rlspinn,
  author    = {Yogatama, Dani and Blunsom, Phil and Dyer, Chris and Grefenstette, Edward and Ling, Wang},
  title     = {Learning to Compose Words into Sentences with Reinforcement Learning},
  booktitle = {International Conference on Learning Representations (ICLR)},
  year      = {2017}
}

@inproceedings{lieck2021rbn,
  author    = {Lieck, Robert and Rohrmeier, Martin},
  title     = {Recursive {Bayesian} Networks: Generalising and Unifying Probabilistic Context-Free Grammars and Dynamic {Bayesian} Networks},
  booktitle = {Advances in Neural Information Processing Systems (NeurIPS)},
  volume    = {34},
  year      = {2021}
}

@article{younger1967cyk,
  author  = {Younger, Daniel H.},
  title   = {Recognition and Parsing of Context-Free Languages in Time $n^3$},
  journal = {Information and Control},
  volume  = {10},
  number  = {2},
  year    = {1967},
  pages   = {189--208}
}

@inproceedings{yang2021pcfg,
  author    = {Yang, Songlin and Zhao, Yanpeng and Tu, Kewei},
  title     = {{PCFGs} Can Do Better: Inducing Probabilistic Context-Free Grammars with Many Symbols},
  booktitle = {Proceedings of NAACL-HLT},
  year      = {2021},
  pages     = {1487--1498}
}

@inproceedings{tran2024pitome,
  author    = {Tran, Hoai-Chau and Nguyen, Duy M. H. and Nguyen, Duy M. and Nguyen, Trung-Tin and Le, Ngan and Xie, Pengtao and Sonntag, Daniel and Zou, James Y. and Nguyen, Binh T. and Niepert, Mathias},
  title     = {Accelerating Transformers with Spectrum-Preserving Token Merging},
  booktitle = {Advances in Neural Information Processing Systems (NeurIPS)},
  year      = {2024}
}

@inproceedings{haurum2024agglomerative,
  author    = {Haurum, Joakim Bruslund and Escalera, Sergio and Taylor, Graham W. and Moeslund, Thomas B.},
  title     = {Agglomerative Token Clustering},
  booktitle = {European Conference on Computer Vision (ECCV)},
  year      = {2024},
  pages     = {200--218}
}

@article{hwang2025dynamic,
  author  = {Hwang, Sukjun and Wang, Brandon and Gu, Albert},
  title   = {Dynamic Chunking for End-to-End Hierarchical Sequence Modeling},
  journal = {arXiv preprint arXiv:2507.07955},
  year    = {2025}
}

@inproceedings{haroun2024dynamic,
  author    = {Haroun, Karim and Allenet, Thibault and Ben Chehida, Karim and Martinet, Jean},
  title     = {Dynamic Hierarchical Token Merging for Vision Transformers},
  booktitle = {Proceedings of the 20th International Joint Conference on Computer Vision, Imaging and Computer Graphics Theory and Applications (VISIGRAPP): Volume 3, VISAPP},
  year      = {2025}
}
\bibliographystyle{iclr2027_conference}

\appendix
\crefalias{section}{appsec}
\clearpage

\section{Model architecture}
\label{app:arch}

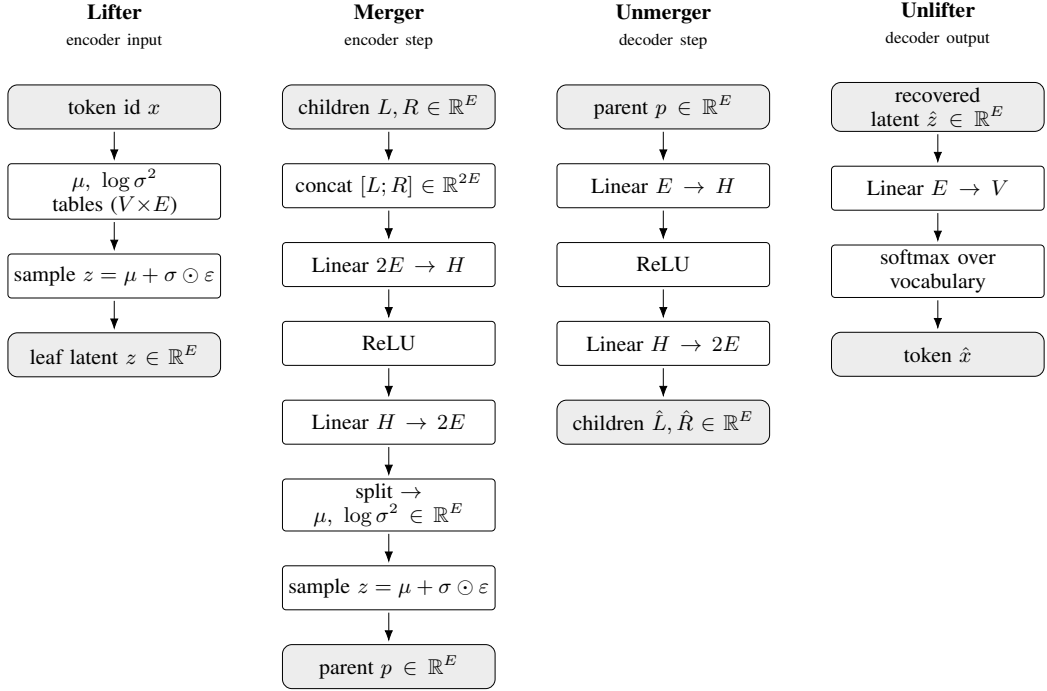
\begin{figure}[!h]
\centering
\resizebox{\textwidth}{!}{\begin{tikzpicture}[
  >=Latex, font=\footnotesize,
  lyr/.style={draw, rounded corners=1.5pt, align=center, text width=3.0cm, minimum height=6.5mm, inner sep=2pt},
  io/.style={draw, rounded corners=4pt, fill=black!7, align=center, text width=3.0cm, minimum height=6.5mm, inner sep=2pt},
  ttl/.style={align=center, text width=3.2cm, font=\small\bfseries},
  ar/.style={->, shorten >=1pt, shorten <=1pt}
]
\node[io] (l0) {token id $x$};
\node[io, right=9mm of l0] (m0) {children $L,R \in \mathbb{R}^{E}$};
\node[io, right=9mm of m0] (u0) {parent $p \in \mathbb{R}^{E}$};
\node[io, right=9mm of u0] (n0) {recovered\\latent $\hat{z} \in \mathbb{R}^{E}$};
\node[ttl, above=4mm of l0] {Lifter\\[1pt]{\scriptsize\mdseries encoder input}};
\node[ttl, above=4mm of m0] {Merger\\[1pt]{\scriptsize\mdseries encoder step}};
\node[ttl, above=4mm of u0] {Unmerger\\[1pt]{\scriptsize\mdseries decoder step}};
\node[ttl, above=4mm of n0] {Unlifter\\[1pt]{\scriptsize\mdseries decoder output}};
\node[lyr, below=5mm of l0] (l1) {$\mu,\ \log\sigma^2$\\tables ($V{\times}E$)};
\node[lyr, below=5mm of l1] (l2) {sample $z=\mu+\sigma\odot\varepsilon$};
\node[io,  below=5mm of l2] (l3) {leaf latent $z \in \mathbb{R}^{E}$};
\draw[ar](l0)--(l1); \draw[ar](l1)--(l2); \draw[ar](l2)--(l3);
\node[lyr, below=5mm of m0] (m1) {concat $[L;R] \in \mathbb{R}^{2E}$};
\node[lyr, below=5mm of m1] (m2) {Linear $2E \to H$};
\node[lyr, below=5mm of m2] (m3) {ReLU};
\node[lyr, below=5mm of m3] (m4) {Linear $H \to 2E$};
\node[lyr, below=5mm of m4] (m5) {split $\to \mu,\ \log\sigma^2 \in \mathbb{R}^{E}$};
\node[lyr, below=5mm of m5] (m6) {sample $z=\mu+\sigma\odot\varepsilon$};
\node[io,  below=5mm of m6] (m7) {parent $p \in \mathbb{R}^{E}$};
\draw[ar](m0)--(m1); \draw[ar](m1)--(m2); \draw[ar](m2)--(m3); \draw[ar](m3)--(m4);
\draw[ar](m4)--(m5); \draw[ar](m5)--(m6); \draw[ar](m6)--(m7);
\node[lyr, below=5mm of u0] (u1) {Linear $E \to H$};
\node[lyr, below=5mm of u1] (u2) {ReLU};
\node[lyr, below=5mm of u2] (u3) {Linear $H \to 2E$};
\node[io,  below=5mm of u3] (u4) {children $\hat{L},\hat{R} \in \mathbb{R}^{E}$};
\draw[ar](u0)--(u1); \draw[ar](u1)--(u2); \draw[ar](u2)--(u3); \draw[ar](u3)--(u4);
\node[lyr, below=5mm of n0] (n1) {Linear $E \to V$};
\node[lyr, below=5mm of n1] (n2) {softmax over vocabulary};
\node[io,  below=5mm of n2] (n3) {token $\hat{x}$};
\draw[ar](n0)--(n1); \draw[ar](n1)--(n2); \draw[ar](n2)--(n3);
\end{tikzpicture}}
\caption{The four networks of the base model, with layers and dimensions. The lifter maps each token id to a
leaf latent through two embedding tables (a mean and a log-variance) and the reparameterisation trick. The
merger is the encoder step, a two-layer perceptron with a ReLU that combines two child latents into a parent;
it is applied bottom-up at every internal node of the tree. The unmerger is the decoder step and an approximate inverse
of the merger, predicting the two children from a parent, applied top-down; its squared error against the true
children is the one-step energy that also scores splits in the chart. The unlifter is a single linear layer
that reads a recovered leaf latent back to a distribution over the vocabulary. A single merger and a single
unmerger are shared across all tree depths.}
\label{fig:arch}
\end{figure}

\Cref{fig:arch} details the four networks of the base model layer by layer. Throughout, $E$ is the
embedding dimension (32 for the base models, 48 for the student), $H$ the hidden dimension (128), and $V$ the
vocabulary size.

\end{document}